# BEnchmarking LLMs for Ophthalmology (BELO) for Ophthalmological Knowledge and Reasoning

Sahana Srinivasan[1,2]*, Xuguang Ai[3]*, Thaddaeus Wai Soon Lo[4], Aidan Gilson[3,5], Minjie Zou[1,2], Ke Zou[1,2], Hyunjae Kim[3], Mingjia Yang[6], Krithi Pushpanathan[1,2], Samantha Yew[1,2], Wan Ting Loke[1,2], Jocelyn Goh[1,2,4], Yibing Chen[4], Yiming Kong[3], Emily Yuelei Fu[3], Michelle Ongyong Hui[7], Kristen Nwanyanwu[8], Amisha Dave[8], Kelvin Zhenghao Li[9], Chen-Hsin Sun[10], Mark Chia[11], Gabriel Dawei Yang[4], Wendy Meihua Wong[1,2,10], David Ziyou Chen[1,2,10], Dianbo Liu[1,2], Maxwell Singer[3,8], Fares Antaki[12,13], Lucian V Del Priore[8], Jost Jonas[4], Ron Adelman[8,14], Qingyu Chen[3]**, Yih-Chung Tham[1,2,4,15]**

1. Centre for Innovation and Precision Eye Health, Department of Ophthalmology, Yong Loo Lin School of Medicine, National University of Singapore, Singapore.
2. Department of Ophthalmology, Yong Loo Lin School of Medicine, National University of Singapore, Singapore.
3. Department of Biomedical Informatics and Data Science, Yale School of Medicine, Yale University, New Haven, USA.
4. Singapore Eye Research Institute, Singapore National Eye Centre, Singapore.
5. Department of Ophthalmology, Massachusetts Eye and Ear, Harvard Medical School, Boston, Massachusetts.
6. School of Chemistry, Chemical Engineering and Biotechnology, Nanyang Technological University, Singapore.
7. Singapore University of Technology and Design, Singapore

8. Department of Ophthalmology and Visual Science, Yale School of Medicine, Yale University, New Haven, USA.

9. Department of Ophthalmology, Tan Tock Seng Hospital, Singapore.

10. Department of Ophthalmology, National University Hospital, Singapore.

11. Centre for Eye Research Australia, Melbourne, Singapore.

12. Cole Eye Institute, Cleveland Clinic, Cleveland, Ohio, USA.

13. The CHUM School of Artificial Intelligence in Healthcare, Montreal, QC, Canada.

14. Department of Ophthalmology, Mayo Clinic, Florida, USA.

15. Eye Academic Clinical Program (Eye ACP), Duke NUS Medical School, Singapore.

*Contributed equally as first authors

**Contributed equally as last authors

## Abstract

**Importance**

Current benchmarks evaluating large language models (LLMs) in ophthalmology are limited in scope and disproportionately prioritise accuracy. We introduce BELO (BEnchmarking LLMs for Ophthalmology), a standardized and comprehensive evaluation benchmark developed through multiple rounds of expert checking by 13 ophthalmologists. BELO assesses ophthalmology-related clinical accuracy and reasoning quality.

**Design and setting:**

Using keyword matching and a fine-tuned PubMedBERT model, we curated ophthalmology-specific multiple-choice-questions (MCQs) from diverse medical datasets (BCSC, MedMCQA, MedQA, BioASQ, and PubMedQA). The dataset underwent multiple rounds of expert checking. Duplicate and substandard questions were systematically removed. Ten ophthalmologists refined the explanations of each MCQ's correct answer. This was further adjudicated by three senior ophthalmologists.

**Main outcomes and measures:**

To illustrate BELO's utility, we evaluated six LLMs (OpenAI o1, o3-mini, GPT-4o, DeepSeek-R1, Llama-3-8B, and Gemini 1.5 Pro) using accuracy, macro-F1, and five text-generation metrics (ROUGE-L, BERTScore, BARTScore, METEOR, and AlignScore). In a further evaluation involving human experts, two ophthalmologists qualitatively reviewed 50 randomly selected outputs for accuracy, comprehensiveness, and completeness.

**Results:**

BELO consists of 900 high-quality, expert-reviewed questions aggregated from five sources: BCSC (260), BioASQ (10), MedMCQA (572), MedQA (40), and PubMedQA (18). To demonstrate BELO's utility, we conducted a series of benchmarking exercises. In the quantitative evaluation, OpenAI's o1 model achieved the highest accuracy (0.88, 95% CI: 0.861–0.903) and macro-F1 score (0.78, 95% CI: 0.869–0.910). On the other hand, the models' performance on text-generation metrics varied and were generally suboptimal, with scores ranging from 20.40 to 71.80 (out of 100, excluding the BARTScore metric), indicating room for improvement in clinical reasoning. In expert evaluations, GPT-4o was rated highest for accuracy and readability, while Gemini 1.5 Pro scored highest for completeness. A public leaderboard has been established to promote transparent evaluation and reporting. Importantly, the BELO dataset will remain a hold-out, evaluation-only benchmark to ensure fair and reproducible comparisons of future models.

**Conclusions and relevance:**

BELO provides a robust clinically relevant benchmark for evaluating both the accuracy and reasoning capabilities of current and emerging LLMs in ophthalmology. Future BELO benchmarking efforts will expand to include vision-based question answering and clinical scenario management tasks.

**Words: 348**

**Introduction**

Large language models (LLMs) have revolutionized the fields of natural language processing (NLP) and Artificial Intelligence (AI). These models, trained on extensive textual data, can respond in a 'human-like' manner to inputs known as prompts.[1] This has catalyzed significant interest in their application in the medical domain[2–5] and its many sub-specialties including ophthalmology.[6–12] LLMs can potentially optimize various tasks including automating clinical documentation, disease diagnosis,[13,14] patient question-answering,[2,10,15] and medical record writing.[16]

Despite these promising benefits, the use of current LLMs in clinical practice comes with risks such as hallucinations[8,10,17] and biased outputs.[18] To improve the accuracy and reliability of LLMs, continuous developments have been made. These include reasoning models that can "think" before answering, such as OpenAI o1 and DeepSeek-R1, which could be better suited for medicine.[10,12] With multiple current and emerging LLMs, both general and medical specific,[19] there is a need for rigorous benchmarking to identify models with the best potential to assist in clinical tasks.[20]

While multiple benchmark datasets currently exist for evaluating LLMs, few are tailored to the ophthalmology domain, which has distinct clinical and linguistic features. For example, the Massive Multitask Language Understanding[21] (MMLU) dataset is widely used by organizations such as OpenAI, Meta, Google and DeepSeek to benchmark their language models across the humanities and STEM. Moreover, several domain-specific benchmarks have emerged such as HealthBench[22] for medicine overall, PathVQA[23] for pathology, and MIMIC-CXR[24] for radiology. Some medical question-answer (QA) datasets, such as MedQA[25] do contain ophthalmological QAs. However, they are not suitable for focused evaluation in ophthalmology due to the lack of metadata to reliably isolate ophthalmology-specific QAs. Furthermore, such datasets often lack gold-standard justifications for the correct answer, which can be used to assess the explanations provided by LLMs for the final answer. Recent efforts, such as Eyecare-Bench[26] and Multi-OphthaLingua,[27] have begun to explore this space, but they primarily evaluate model accuracy alone. Currently ophthalmological evaluation studies also use various, non-standardized validation datasets as seen in Supplementary Table 1.[12,28–32] This hinders the ability to track and compare the performance of different LLMs. To date, there is no standardized benchmark that rigorously tests both ophthalmic knowledge and the reasoning behind the answer in a structured, reproducible manner.

To address this critical gap, we establish BELO (BEnchmarking LLMs for Ophthalmology), a comprehensive benchmark dataset evaluating LLMs on ophthalmological knowledge and clinical reasoning. BELO combines and augments data from diverse medical datasets with multiple rounds of expert checking. This study represents the first phase of a broader initiative to develop ophthalmology-specific benchmarks, establishing a rigorous and clinically relevant framework to evaluate current and emerging LLMs in Ophthalmology.

**Methods**

Overview of BELO

**Figure 1** provides an overview of the dataset curation and grading workflow, while **Figure 2** illustrates the dataset filtering and selection process. In summary, we first extracted ophthalmological QAs from multiple-choice question (MCQ) datasets that test medical knowledge using a combination of keyword matching and fine-tuned PubMedBERT model-based approaches. The dataset was quality-checked through rigorous multi-stage expert checking and questions with substandard reasoning were manually refined. Approval from the ethics committee was not required since patients were not involved in our study. The study followed the Strengthening the Reporting of Observational Studies in Epidemiology (STROBE) reporting guideline.

Collection of medical datasets

We systematically reviewed and collected existing medical QA datasets (**Figures 1 and 2**). The search used multiple platforms, including Hugging Face, Google Scholar, PubMed Central, Kaggle, and Papers with Code. Only English-language datasets were considered. Additional details can be found in the **Supplementary Materials' Appendix** section. This comprehensive search strategy initially yielded 33 datasets (**Figure 2**). However, 27 were excluded as their QA was not in MCQ format, and one was further excluded due to its closed-sourced nature. Ultimately, five MCQ datasets, Basic and Clinical Science Course (BCSC) (n = 3,542), BioASQ Task B Phase B[33] (n = 5,109), MedMCQA[34] (n = 187,005),

MedQA[25] (n = 12,723), and PubMedQA[35] (n = 1,000), were selected (**Table 1 and Figure 2**). Detailed information about these datasets can be found in the **Supplementary Materials' Appendix** section. These datasets were then standardized and formatted into JSON files with uniform file structures to ensure consistency in data representation.

Ophthalmological QA extraction

Subsequently, ophthalmological QA pairs from these datasets were extracted using two approaches: keyword matching and a finetuned PubMedBERT-based[36] approach for BioASQ, MedQA and PubMedQA, and de-duplicated as seen in **Figure 1**. Both methods were evaluated on the full MedMCQA dataset to assess their efficacy. The detailed descriptions of the keyword matching and finetuned PubMedBERT-based approaches can be found in the **Appendix**. For MedMCQA, we leveraged the dataset's predefined ophthalmology subject labels to extract relevant QA pairs (**Figure 1**). As BCSC is an ophthalmological dataset, no extractions were needed.

We performed stratified random sampling from the ophthalmological QAs extracted from each of the five source datasets, selecting items in proportion to each dataset's total ophthalmological question count, so that the initial pool totaled 900 questions as seen in **Figure 2**. This number was selected to balance content diversity with the practical constraints of expert-driven checking and reasoning refinement. However, because BioASQ, MedQA, and PubMedQA each contributed fewer than 5% of the total ophthalmology items, their proportional representation was adjusted by oversampling additional QAs. This step ensured that the final BELO benchmark was not disproportionately dominated by questions from MedMCQA.

QA Quality Check

Manual Quality Check

All 900 QA items were thoroughly checked by one board-certified ophthalmologist (MZ), two optometrists (SY, WTL) and six research staff (SS, XA, TWSL, SY, WTL, YC) to remove non-ophthalmological QAs (false positives) and duplicate questions as seen in **Figures 1 and 2**. These items was replaced by randomly chosen ophthalmological questions from the same dataset to maintain proportional representation of the dataset. Items were then graded based on the quality of their QA as well as the presence of reasonings. Items were graded one if the item had either no reasonings or bad quality reasonings. We classified a reasoning as poor quality if it met any of the following criteria (adapted from prior work).[12] Firstly, the reasoning was considered poor quality if it consisted solely of isolated terms or phrases without forming a coherent explanation. The second criterion was if the reasoning simply restated the correct option (or enumerated all options) without providing justification. The third criterion to be considered poor quality was if it failed to explain why the correct choice was correct. Additionally, the items were graded one if the question, options or correct answer was outdated or wrong. Items were graded two if the reasoning explained the correct option alone and graded three if the reasoning explained both the correct option and the incorrect options correctly.

Reasoning Amendment Exercise

The items graded two or three were added into the BELO dataset as-is. The items graded one were presented to 10 board-certified ophthalmologists from Singapore (DC, WW, CHS, MZ), Australia (MC), Canada (FA) and the USA (AG, MS, KNH, AD) to improve (**Figures 1 and 2**). Each QA item was revised by one board-certified ophthalmologist who was tasked to clearly justify the correct answer and, when informative, to explain why the incorrect options were wrong. Items identified to be outdated were replaced with other questions from the same dataset, deemed to be up to date and verified again by the ophthalmologist. Items with either ill-phrased questions, or incorrect options and suggested correct answers, were either corrected or replaced with questions from the same dataset. The replaced or corrected questions were rechecked by the ophthalmologist who flagged them, to ensure accuracy and quality. After the amendments, the question items were proof-read, further amended and adjudicated by three senior ophthalmologists from the USA (RA, LDP) and Germany (JJ) (**Figure 1**).

Benchmarking LLMs for the BELO Leaderboard

As a demonstration of BELO, six models were evaluated on the BELO dataset: OpenAI o1 (o1-preview-2024-09-12), OpenAI o3-mini (o3-mini-2025-01-31), GPT-4o (gpt-4o-2024-05-13), DeepSeek-R1, Llama-3-8B (Meta-Llama-3-8B-Instruct), and Gemini 1.5 Pro (gemini-1.5-pro-001). All six models were accessed via their respective application programming interface (API).

BELO Leaderboard Website

The results of the quantitative evaluations have been published on a public leaderboard to facilitate standardized comparisons across models. This serves as a reference point for future studies evaluating performance on the BELO benchmark. The link to the website is as follows: https://belo-dataset.vercel.app/.

Prompt Engineering

As prompt phrasing has been noted to influence model performance significantly,[12,37] the input prompt style was standardized and aligned to styles used in previous studies.[12,38] An example of the prompt used can be found in Supplementary Figure 1. The prompt included the context, instruction, input question item, as well as the output data format. The models were tested in a zero-shot manner and were tasked to output the correct option answer to the MCQ question as well as the reasoning behind the answer in a JSON format.

Demonstrative Quantitative Analysis Using BELO Across Six Models

The outputs of the six models were quantitatively evaluated in terms of accuracy and model reasoning as seen in **Figure 1**. Model accuracy was assessed as (1) accuracy and through (2) macro-F1, which is the unweighted harmonic mean of recall and precision.[12] To ensure fair comparison across models, macro-F1 was computed using only questions with four options, which were questions from BCSC and MedMCQA. As BioASQ includes two answer options and PubMedQA includes three answer options, they were not included to prevent distortion of class balance. Model reasoning was evaluated using five text-generation metrics used in prior studies, Recall-Oriented Understudy for Gisting Evaluation (ROUGE-L),[39] BERTScore,[40] BARTScore,[41] AlignScore[42] and the Metric for Evaluation of Translation with Explicit Ordering (METEOR).[12,43–47] They evaluated the model output compared to the ground truth based on a variety of methods. The definitions of the five text-generation metrics can be found in the **Appendix.** For all metrics, a higher score indicates better performance, while for the negatively valued BARTScore, a score closer to 0 indicated better performance.[12]

To quantify overall reasoning quality, we computed a weighed normalized score for each model by first rescaling each text-generation metric to the [0,1] interval, assigning 0 to the lowest observed value and 1 to the highest, with linear interpolation for intermediate results. We then averaged these five normalized metric scores with equal weights to yield a single aggregate reasoning performance score per model.

Demonstrative Qualitative Evaluation Using BELO Across Three Representative Models

LLM evaluations need to include qualitative evaluations to complement quantitative evaluations. Therefore, we conducted demonstrative qualitative evaluation of the reasoning processes of three representative models, GPT-4o, Llama-3-8B, and Gemini 1.5 Pro. The models were evaluated by two board-certified ophthalmologists (FA, KL) based on accuracy, completeness, and readability. This evaluation was conducted on 50 randomly selected questions for which all three models had produced the correct answer. The ophthalmologists rated each model response using a 5-point Likert scale. The rubrics for evaluation can be found in Supplementary Table 2.

Statistics

All analyses were performed in Python (v3.9.0, Python Software Foundation) using two-sided tests throughout. To compare model accuracy, we applied independent two-sample t-tests, and for macro-F1 scores we used z-tests. Although our data is non-parametric, the large aggregate sample size (n = 900) permits the central limit theorem to justify these parametric tests. We also employed Bonferroni adjustment to maintain the family-wise error rate across multiple comparisons.

For all text-generation metrics, we compared the distributions of metric scores between models using two-tailed Wilcoxon rank-sum tests, selected because the metric distributions are non-normal and all models were evaluated on BELO independently of each other. Bonferroni correction was again applied to account for multiple pairwise tests. All P values shown are Bonferroni corrected.

**Results**

**Finalized Benchmarking Dataset**

To extract ophthalmology-related questions from MedMCQA, we used a combination of fine-tuned PubMedBERT classifier and a rule-based keyword matching method. The original MedMCQA has 6,990 ophthalmological and 187,015 non-ophthalmological questions as the ground truth. The keyword matching method identified 9,825 items as ophthalmological, of which 5,016 were correctly identified and with a sensitivity of 0.718 (Supplementary Table 3a). This methodology identified 177,180 items as non-ophthalmological, out of which 175,206 were correctly identified with a specificity of 0.973. The finetuned PubMedBERT model identified 13,407 questions as ophthalmological, out of which 6,551 were correctly identified with a sensitivity of 0.937 (Supplementary Table 3b). This methodology identified 173,598 items as non-ophthalmological, out of which 173,159 were correctly identified with a specificity of 0.962.

Eventually, we used a combined strategy using both the keyword and finetuned PubMedBERT methods. MedMCQA had the highest number of ophthalmological QAs (n=6,990), followed by BCSC (n=3542), MedQA (n=429), PubMedQA (n= 26) and BioASQ (n=21). QA items were then randomly selected from each dataset with the final curated BELO dataset consisting of 900 questions from MedMCQA (n=572), BCSC (n=260), MedQA (n=40), PubMedQA (n=18) and BioASQ (n=10).

Results of the Manual Quality Check

Out of the 900 QA items, 510 question items were flagged as either non-ophthalmological or duplicates and replaced with other ophthalmological QA items from the same dataset. In the graded manual quality check, 40 items (4.4%) had no reasoning, 92 items (10.2%) were labeled as 1 (low-quality), 382 items (42.4%) were labeled as 2 (explained the correct answer alone) and 386 items (42.9%) were labeled as 3 (explained both the correct and incorrect options) (Supplementary Table 4). Notably, all 40 items without reasoning were from MedQA and all 92 items labeled as 1 were from MedMCQA.

Quantitative analysis

Demonstrative Quantitative Evaluations - Accuracy and Macro-F1 Score

Table 2 and Figure 3A highlight the LLMs' performance in accuracy macro-F1. Overall, OpenAI o1 demonstrated the best performance in accuracy (0.882; 95% CI: 0.861 - 0.903), followed by DeepSeek-R1 (0.876; 95% CI: 0.854 - 0.898), OpenAI o3-mini (0.856; 95% CI: 0.833 - 0.879), GPT-4o (0.831; 95% CI: 0.807 - 0.855), Llama-3-8B (0.741; 95% CI: 0.712 - 0.770) and Gemini 1.5 Pro (0.596; 95% CI: 0.564 - 0.628). There were no statistically significant differences among o1, o3-mini and DeepSeek-R1. Although GPT-4o did not differ significantly in accuracy from o3-mini and DeepSeek-R1, it performed significantly worse than o1 ($P$ = 0.0294). All four top performing models (o1, o3-mini, DeepSeek-R1 and GPT-4o) outperformed Llama-3-8B and Gemini 1.5 Pro with high statistical significance ($P$ < 0.001).

A similar trend was found for macro-F1 (Table 2 and Figure 3B). OpenAI o1 demonstrated the best performance in macro-F1 (0.890; 95% CI: 0.869 - 0.910), followed by DeepSeek-R1 (0.884; 95% CI: 0.863 - 0.905), OpenAI o3-mini (0.858; 95% CI: 0.835 - 0.880), GPT-4o (0.835; 95% CI: 0.809 - 0.859), Llama-3-8B (0.744; 95% CI: 0.716 - 0.774) and Gemini 1.5 Pro (0.639; 95% CI: 0.606 - 0.672) ($P > 0.99$ between DeepSeek-R1 and o1, $P = 0.00186$ between o3-mini and DeepSeek-R1, $P = 0.0172$ between o3-mini and GPT-4o, $P < 0.001$ between all other comparisons).

Demonstrative Quantitative Evaluations - Text-Generation Metrics

Table 2 and Figures 4A-E highlight the models' performances in the five text generation metrics. In summary, based on the weighted normalized metric, OpenAI o1 performed the best, followed by o3-mini, GPT-4o, DeepSeek-R1, Llama-3-8B and Gemini 1.5 Pro.

In ROUGE-L, GPT-4o (0.204; 95% CI: 0.200 - 0.209) emerged joint-best with o3-mini (0.203; 95% CI: 0.199 - 0.207) (Figure 4A). There was no statistical difference between these two models ($P > 0.99$ between GPT-4o and o3-mini, $P < 0.001$ compared to the rest of the models). This was followed by Llama-3-8B (0.187; 95% CI: 0.183 - 0.191), o1 (0.186; 95% CI: 0.183 - 0.190) and DeepSeek-R1 (0.186; 95% CI: 0.182 - 0.190) and Gemini 1.5 Pro (0.163; 95% CI: 0.159 - 0.167).

In METEOR, o1 ranked first (0.247; 95% CI: 0.243 - 0.253), followed by o3-mini (0.228; 95% CI: 0.222 - 0.234), GPT-4o (0.227; 95% CI:.221 - 0.233), DeepSeek-R1 (0.225; 95% CI: 0.220 - 0.231), Gemini 1.5 Pro (0.225; 95% CI: 0.219 - 0.229) and Llama-3-8B (0.224; 95% CI: 0.220 - 0.230) ($P < 0.001$ between o1 and the rest of the models) (Figure 4B).

In BERTScore, o3-mini (0.718; 95% CI: 0.715 - 0.720) ranked joint-first with GPT-4o (0.713; 95% CI: 0.711 - 0.716) (Figure 4C). This was followed by o1 (0.712; 95% CI: 0.710 - 0.714), DeepSeek-R1 (0.711; 95% CI: 0.709 - 0.713), Llama-3-8B (0.703; 95% CI: 0.700 - 0.705) and Gemini-1.5-Pro (0.690; 95% CI: 0.688 - 0.692) ($P < 0.05$ comparing o3-mini to the rest, except $P = 0.665$ between o3-mini and GPT-4o).

In BARTScore, o1 (-3.289; 95% CI: -3.335 - -3.241) ranked joint-best with DeepSeek-R1 (-3.313; 95% CI: -3.363 - -3.264) and o3-mini (-3.360; 95% CI: -3.409 - -3.313) ($P > 0.99$ for all comparisons between these 3 models except $P = 0.590$ between o1 and o3-mini) (Figure 4D). This was followed by GPT-4o (-3.606; 95% CI: -3.663 - -3.549), Gemini 1.5 Pro (-3.628; 95% CI: -3.682 - -3.574) and Llama-3-8B (-3.683; 95% CI: -3.737 - -3.630) ($P < 0.001$ when comparing GPT-4o, Gemini 1.5 Pro and Llama-3-8B with o1, DeepSeek-R1 and o3-mini).

Finally, in AlignScore, GPT-4o (0.255; 95% CI: 0.244 - 0.265) ranked joint-first with o3-mini (0.252; 95% CI: 0.241 - 0.262) and o1 (0.236; 95% CI: 0.228 - 0.245) with no

significant difference among these three models (all $P > 0.99$ except $P = 0.931$ between GPT-4o and o1) (Figure 4E). These three models significantly outperformed DeepSeek-R1 (0.209; 95% CI: 0.201 - 0.217), Llama-3-8B (0.201; 95% CI: 0.192 - 0.209), and Gemini 1.5 Pro (0.197; 95% CI: 0.189 - 0.204). ($P < 0.001$). Among DeepSeek-R1, Llama-3-8B, and Gemini 1.5 Pro, no significant differences were observed ($P > 0.99$).

Qualitative Analysis

In terms of accuracy, while GPT-4o ranked first (4.91), it performed comparably with Gemini 1.5 Pro (4.81) and Llama-3-8B (4.72). Similarly, in completeness, while Gemini 1.5 Pro (4.79) ranked first, it performed comparably with GPT-4o (4.68) and Llama-3-8B (4.59). However, in readability, GPT-4o (4.92) performed comparably with Llama-3-8B (4.78) but outperformed Gemini 1.5 Pro (4.71) ($P = 0.00242$ between GPT-4o and Gemini 1.5 Pro) (Supplementary Table 5).

**Discussion**

In this study, we introduce BELO, a robust, comprehensive benchmark designed to evaluate both accuracy and clinical reasoning in ophthalmology. The uniqueness and strengths of BELO include its establishment via multi rounds expert checking, clinical reasoning compositions for all QA pairs, and a transparent leaderboard-style evaluation and reporting. As an illustration on the utility and relevancy of BELO, we evaluated six LLMs quantitatively in terms of accuracy, macro-F1, and five text-generation metrics (ROUGE-L, BERTScore, BARTScore, METEOR, and AlignScore). Furthermore, two ophthalmologists evaluated 50 random outputs of three representative models qualitatively in terms of reasoning accuracy, comprehension and readability. We observed that o1 ranked first in accuracy, macro-F1 and text-generation metrics overall. In the expert qualitative evaluation, GPT-4o was ranked highest in accuracy and readability, while Gemini 1.5 Pro had the highest score in completeness. Overall, BELO is a robust, clinical-relevant framework for the evaluation of LLMs in ophthalmology.

Prior work in ophthalmological benchmarking has focused on visual QA tasks such as classification, segmentation, short-answer QA, and report generation.[26,27] These studies often lacked systematic expert checking of QA items and did not include manually curated reasonings, limiting their effectiveness in evaluating model clinical reasoning.[26,27] Furthermore, most efforts have relied on either private datasets or existing open-source datasets such as MedMCQA,[12,38] while broader medical LLM studies often use MedQA[37] or PubMedQA.[2,8] In contrast, BELO introduces standardized, text-based LLM evaluation in ophthalmology, consolidating ophthalmology-specific questions from diverse medical QA sources, including both ophthalmological examination materials and scientific QA datasets into a unified benchmark.

To ensure the quality and reliability of the benchmark, BELO underwent a four-stage expert checking process. Two optometrists, nine board-certified ophthalmologists and three senior ophthalmologists systematically reviewed and refined all low-quality or outdated parts of question items. This approach ensured that each item met gold-standard criteria, with reasonings that comprehensibly incorporate all relevant and necessary information to determine the correct answer.

The results from BELO's demonstrative qualitative evaluations highlight its discriminative power. Generally, newer models outperformed older models. In terms of both accuracy and macro-F1, OpenAI o1 ranked first, followed by DeepSeek-R1 and o3-mini. However, performance across the text-generation metrics was generally low. O1 still ranked first overall, followed by o3-mini and GPT-4o. This suggests limitations in ophthalmological reasoning, consistent with other benchmarking papers where generally, o1 and GPT-4o perform the best but exhibited limitations in medical reasoning.[10,15,38] To complement quantitative evaluations, we also performed demonstrative qualitative evaluations of the reasonings of GPT-4o, Llama-3-8B, and Gemini 1.5 Pro in terms of accuracy, completeness, and readability. OpenAI o1, o3-mini, and DeepSeek-R1 were excluded due to their release after the qualitative evaluation period. Overall, the three models performed comparably, with GPT-4o performing better than Gemini 1.5 Pro in readability with statistical difference.

BELO offers several strengths as a domain-specific benchmark for ophthalmology. In this first phase of BELO, it includes 900 ophthalmology-focused QA items, each systematically and rigorously checked several rounds by ophthalmologists to ensure ophthalmic clinical relevance and accuracy. Secondly, BELO also provides evaluation of model reasoning with reference standards curated and adjudicated by ophthalmologists. Thirdly, by sourcing and repurposing items from several established QA datasets that encompass diverse examination formats and geographic settings, BELO may allow a broader assessment of clinical knowledge in ophthalmology. Finally, BELO also incorporates a public leaderboard platform (https://belo-dataset.vercel.app/) to support transparent model comparisons and reporting. The BELO dataset will be retained as a hold-out, evaluation-only benchmark to ensure fair and reliable model comparisons.

Despite its strengths, BELO has several limitations that future work should address. Currently, BELO has insufficient questions from real-world clinical cases. To better reflect real-world clinical tasks, future phases of BELO will build upon our existing visual-language benchmark, the Large Multimodal Ophthalmology Dataset (LMOD)[48] to incorporate visual question-answering items that evaluate multimodal reasoning. Furthermore, questions reflecting real-world clinical cases with multiple follow ups and corresponding management plans will be included in our future works to enhance clinical fidelity.

**Conclusion**

BELO is a comprehensive benchmark designed to evaluate LLMs in ophthalmology, in terms of both model accuracy and reasoning. Unlike earlier benchmarks that focus solely on answer correctness, BELO incorporates a rich text corpus paired with expert-curated explanations, serving as a reference standard. BELO also includes a public leaderboard to facilitate transparent performance evaluation and reporting. BELO can potentially serve as a robust and clinically relevant framework for assessing the capabilities of current and emerging LLMs in ophthalmology. Future BELO benchmarking efforts will be expanded to include vision LLMs.

Figures

Figure 1: Overview of the BELO Dataset Curation Pipeline

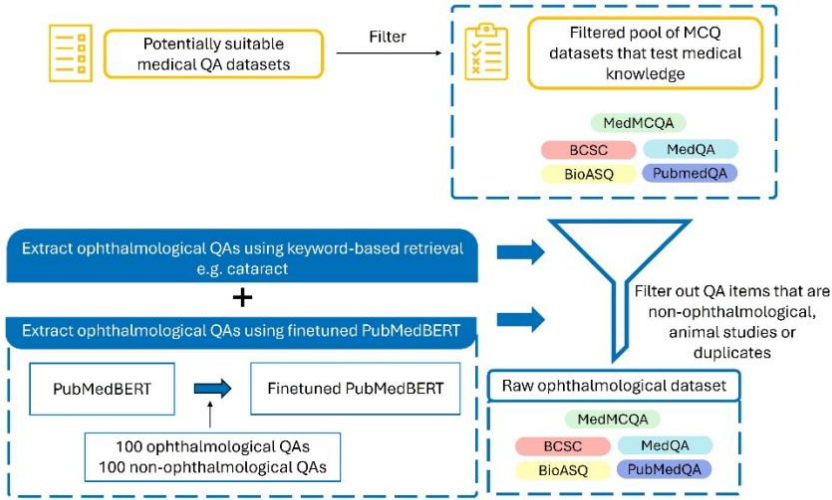

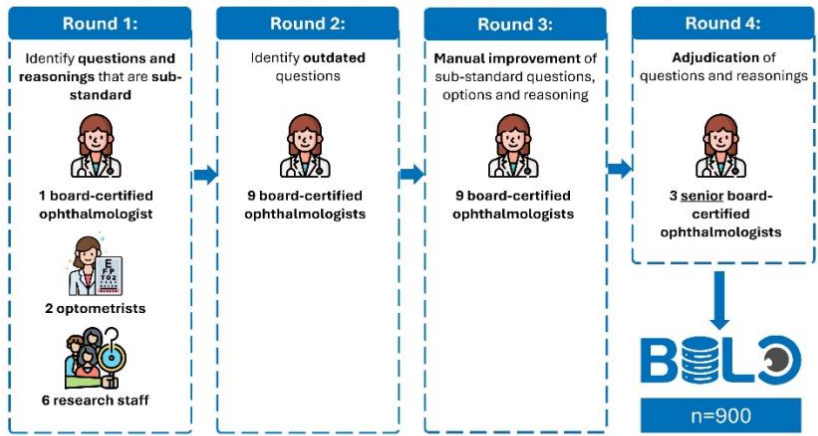

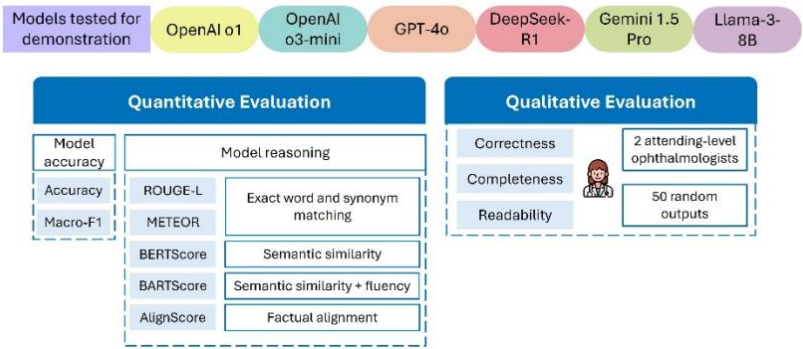

Figure 1 Legend: BELO was developed through a three-step process: (1) collection of medical question answer (QA) datasets and extraction of ophthalmology-specific QA pairs, (2) multi-stage expert checking to ensure the quality and clinical relevance of all QA items, and (3) demonstrative benchmarking of leading LLMs to populate the BELO leaderboard.

Figure 2: End-to-End Pipeline for Building the BELO Benchmark Dataset

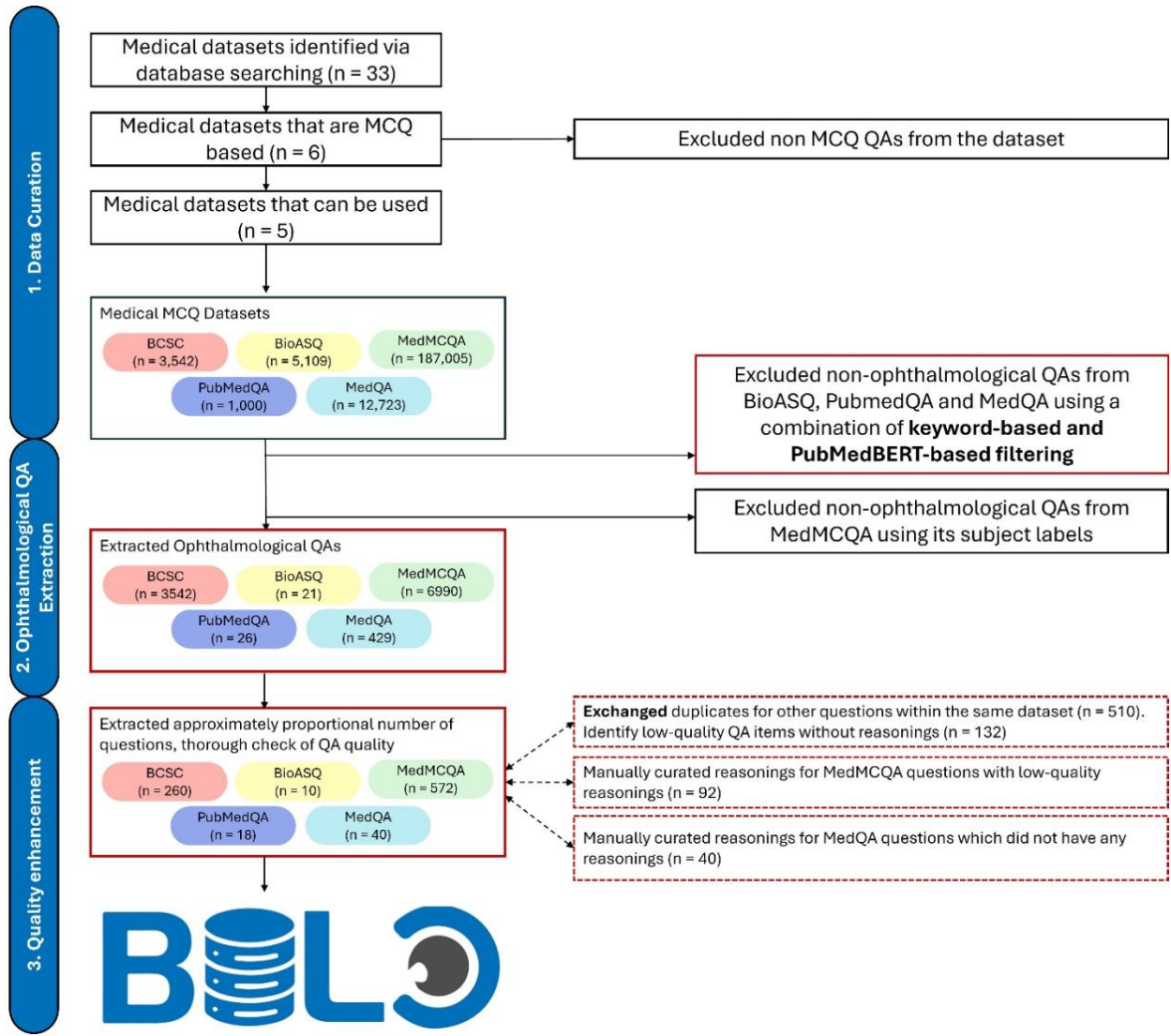

Figure 2 Legend: Five medical datasets were selected from an initial pool of 33 to construct BELO. An approximately proportional number of ophthalmology-related question–answer (QA) pairs were sampled from each dataset using stratified random sampling. For datasets contributing fewer than 5% of the total ophthalmology QA pool—

namely BioASQ, PubMedQA, and MedQA—additional QA pairs were included to ensure adequate representation.

Figure 3: Comparison of Model Performance on BELO using A) Accuracy and B) Macro-F1

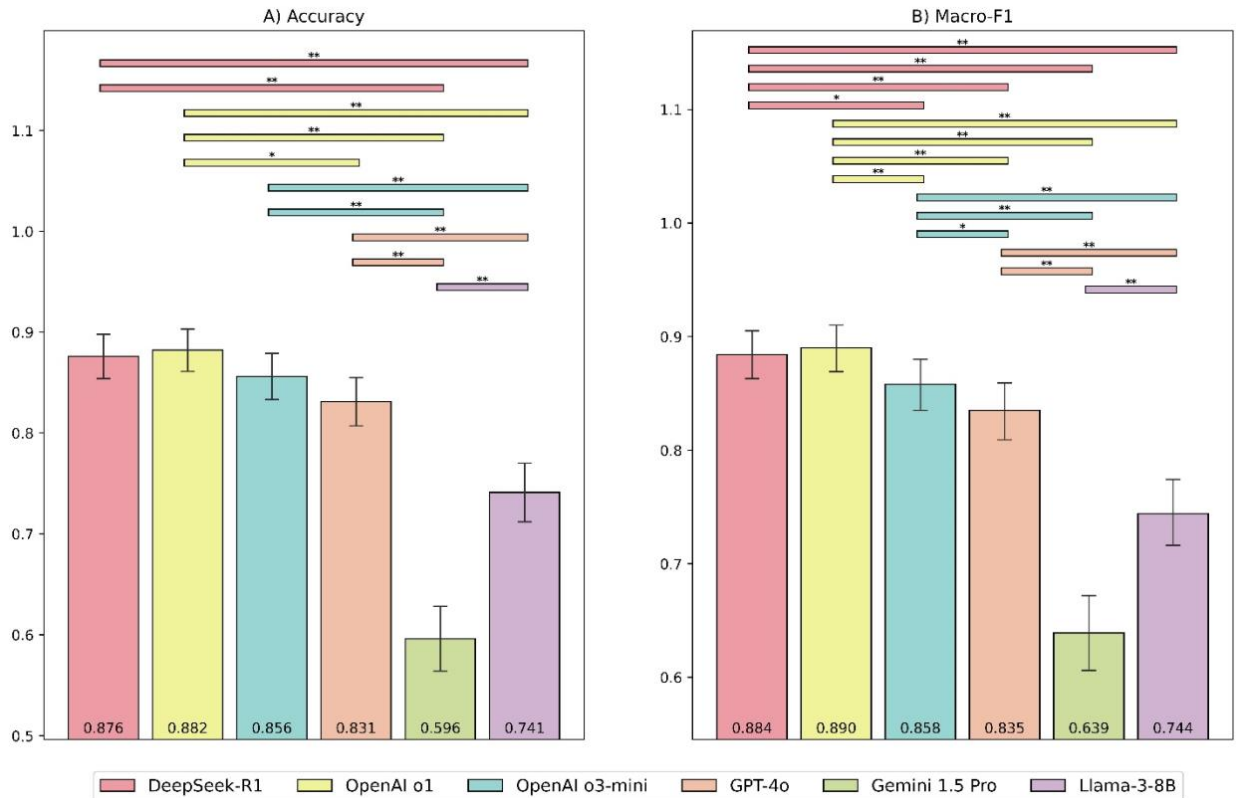

Figure 3 Legend: Bar plots show the mean performance scores with 95% confidence intervals for each model. Horizontal bars above each plot indicate pairwise comparisons; the color of each bar corresponds to the model with significantly higher performance. Accuracy was computed using the full BELO dataset (n = 900). To ensure fair comparison across models, macro-F1 was calculated using only questions from BCSC, MedMCQA, and MedQA, which uniformly contain four answer options, resulting in a subset of 872 questions.

*denotes $P < 0.05$,

**denotes $P < 0.001$

Figure 4: Comparison of 6 LLMs on BELO in terms of 5 text generation metrics - A) ROUGE-L, B) METEOR, C) BERTScore, D) BARTScore, and E) AlignScore

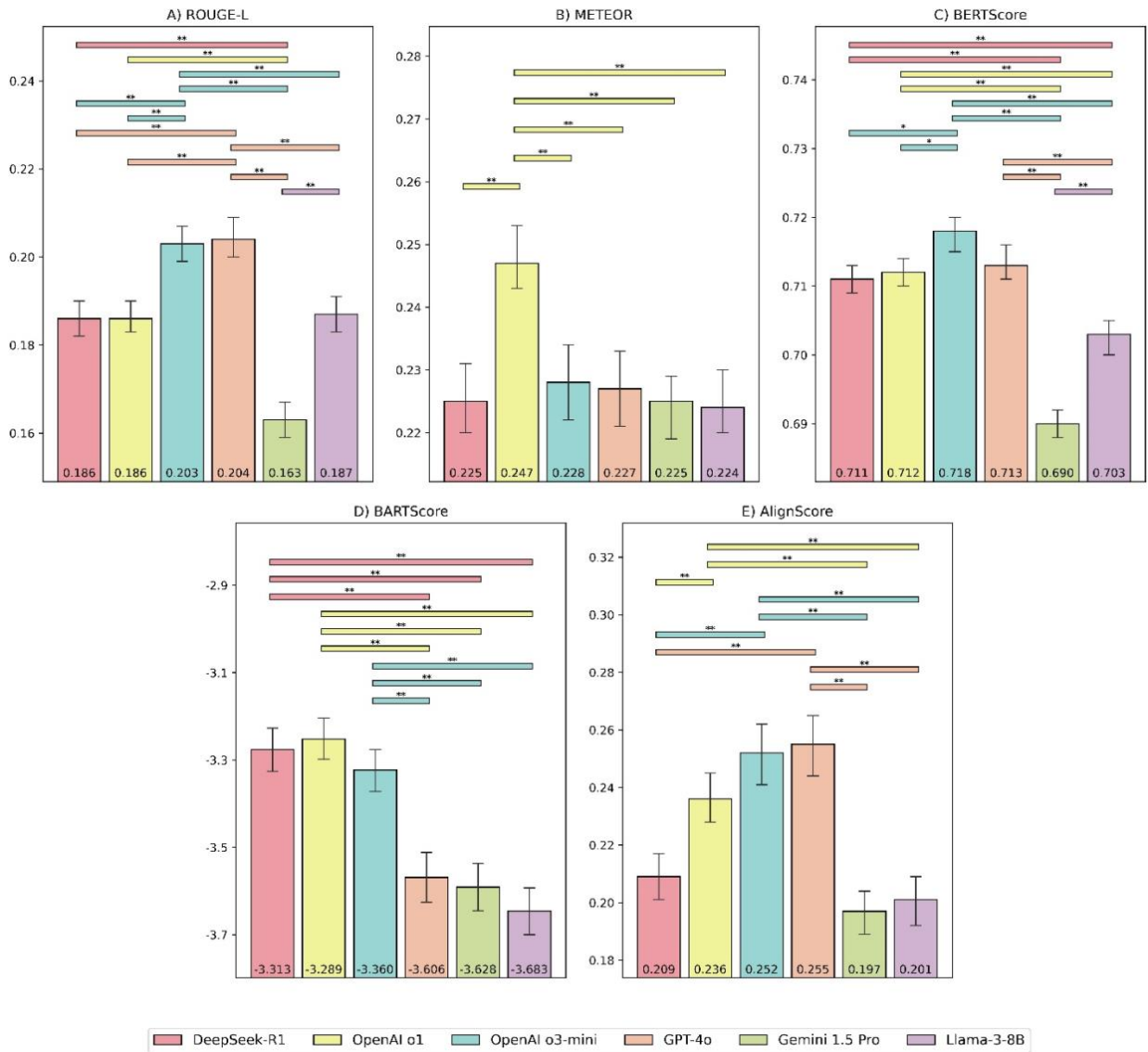

Figure 4 Legend: Bar plots show the mean metric score with 95% confidence interval for each model. Horizontal bars above each plot indicate pairwise comparisons; the color of each bar corresponds to the model with significantly higher performance.

*denotes $P < 0.05$,

**denotes $P < 0.001$

Figure S1: Illustration of the prompt used to test the six models in a standardized manner

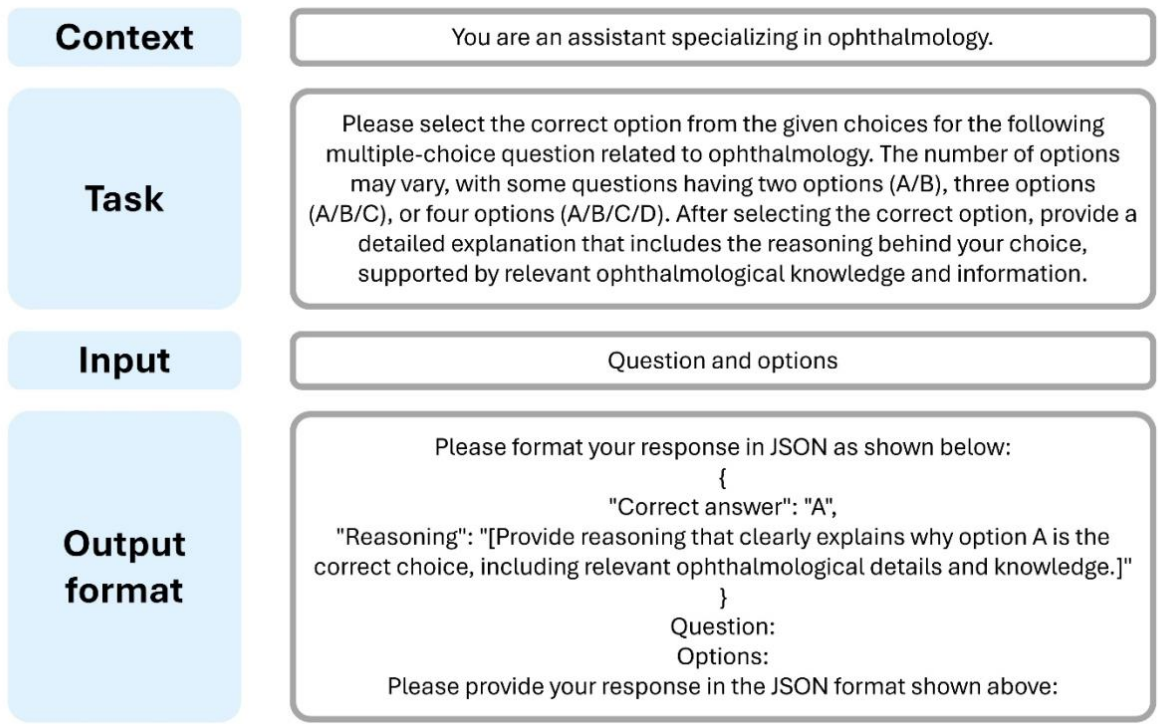

Figure S1 Legend: Each of the 900 questions was formatted into a standardized prompt structure – context, instruction, input question item and output data format - before being used to test the six models in a zero-shot manner.

Table 1: Compositions of the Curated BELO dataset

| Dataset | BCSC | BioASQ Task B Phase B | MedMCQA | MedQA | PubMedQA |
|---|---|---|---|---|---|
| Number of Questions in BELO | 260 | 10 | 572 | 40 | 18 |
| Number of reasonings amended | 0 | 0 | 93 | 40 | 0 |
| Question type | MCQ | Yes/No | MCQ | MCQ | Yes/No/Maybe |
| Number of options | 4 | 2 | 4 | 4 | 3 |
| Task | Ophthalmological examination questions | Biomedical research questions | Indian AIIMS and NEET PG entrance examination questions | USMLE style board examination questions | Biomedical research questions |
| Presence of ophthalmological labels | Fully ophthalmological | No | Yes | No | No |
| Presence of reasoning in original dataset | Present | Present | Present | Absent | Present |

Table 2: Performance the of Six LLMs in the Demonstrative Quantitative Evaluation in answering BELO questions in terms of Accuracy, Macro-F1 and Five Text-Generation Metrics

| Metric | Number of QAs | DeepSeek-R1 | OpenAI o1 | OpenAI o3-mini | GPT-4o | Gemini 1.5 Pro | Llama-3-8b |
|---|---|---|---|---|---|---|---|
| Macro-F1 | 872 | 0.884 (0.863 - 0.905)^ | **0.890 (0.869 - 0.910)** | 0.858 (0.835 - 0.880) | 0.835 (0.809 - 0.859) | 0.639 (0.606 - 0.672) | 0.744 (0.716 - 0.774) |
| Accuracy | 900 | 0.876 (0.854 - 0.898) | **0.882 (0.861 - 0.903)** | 0.856 (0.833 - 0.879) | 0.831 (0.807 - 0.855) | 0.596 (0.564 - 0.628) | 0.741 (0.712 - 0.770) |
| ROUGE-L | | 0.186 (0.182 - 0.190) | 0.186 (0.183 - 0.190) | 0.203 (0.199 - 0.207) | **0.204 (0.200 - 0.209)** | 0.163 (0.159 - 0.167) | 0.187 (0.183 - 0.191) |
| METEOR | | 0.225 (0.220 - 0.231) | **0.247 (0.243 - 0.253)** | 0.228 (0.222 - 0.234) | 0.227 (0.221 - 0.233) | 0.225 (0.219 - 0.229) | 0.224 (0.220 - 0.230) |
| BERTScore | | 0.711 (0.709 - 0.713) | 0.712 (0.710 - 0.714) | **0.718 (0.715 - 0.720)** | 0.713 (0.711 - 0.716) | 0.690 (0.688 - 0.692) | 0.703 (0.700 - 0.705) |
| BARTScore | | -3.313 | **-3.289** | -3.360 | -3.606 | -3.628 | -3.683 |

| | | (-3.363 - -3.264) | **(-3.335 - -3.241)** | (-3.409 - -3.313) | (-3.663 - -3.549) | (-3.682 - -3.574) | (-3.737 - -3.630) |
|---|---|---|---|---|---|---|---|
| AlignScore | | 0.209 (0.201 - 0.217) | 0.236 (0.228 - 0.245) | 0.252 (0.241 - 0.262) | **0.255 (0.244 - 0.265)** | 0.197 (0.189 - 0.204) | 0.201 (0.192 - 0.209) |
| Weighted Normalized Scores* | | 0.500 | **0.804** | 0.784 | 0.629 | 0.037 | 0.224 |

The bolded item refers to the best performance in that metric.

^The scores presented are the means followed by the 95% Confidence Interval Range.

* Model performance in each text generation metric was normalized on a scale from 0 to 1, with 0 being assigned to the lowest observed score and 1 to the greatest, with linear interpolation for intermediate results. The weighted normalized scores were calculated by averaging these five normalized metric scores with equal weights to yield a single aggregate reasoning performance score per model.

Supplementary Materials

Table S1: Overview of Existing Ophthalmological Question-Answer (QA) Datasets Used in Benchmarking Studies and BELO

| Evaluation dataset used | Dataset size | Question type | Presence of reasoning | License |
|---|---|---|---|---|
| MedMCQA | 6990 | MCQ | Present with sub-standard quality | Open source |
| BCSC and Ophthoquestions | 260 | MCQ | Present | Closed source, Licensed |
| OphthalVQA | 600 | Visual short answer question | Absent | Open source |
| Ophthalmology single-choice questions | 100 | MCQ | Absent | Open source but unreleased |
| Multi-OphthaLingua | 1184 | MCQ | Absent | Open source |
| Eyecare-Bench | 15,000 | Visual Question Answering | Absent | Unreleased |
| OphthBench | 591 | MCQ, Open ended questions, short answer questions in Chinese | Absent | Unreleased |
| EyeQA-Plus | 15k | Synthetic open-ended questions | Absent | Open source |
| **BELO** | **900** | **MCQ** | **Present and quality checked** | **Intended exclusively as a held-out validation set** |

Table S2: Rubrics Used to Evaluate GPT-4o, LLaMA-3-8B, and Gemini 1.5 Pro Qualitatively in Terms of Accuracy, Completeness and Readability

| Metric | Likert Scale |
|---|---|
| Accuracy of response: Assess whether the model selected the correct answer and provided accurate reasoning. | Rating ranges from 1 (bad) 2,3,4 to 5 (good): <br><br> 1 (bad): The model chose an incorrect option, and the reasoning is factually incorrect. <br><br> 2: The model chose the correct option, but the reasoning is mostly incorrect or irrelevant. <br><br> 3: The model chose the correct option, but the reasoning is partially inaccurate or incomplete. <br><br> 4: The model chose the correct option, with mostly accurate reasoning but minor errors. <br><br> 5 (good): The model chose the correct option with fully accurate reasoning. <br><br> Additional annotations: if the response contains false or misleading information, please identify them. |
| Completeness of Response: Assess whether the reasoning provided by the model is comprehensive and captures all necessary aspects to justify the chosen answer. | 1 (bad): The reasoning is incomplete, missing key information or crucial details needed to justify the answer.<br><br>2: The reasoning is somewhat complete, but it lacks key information that impacts its comprehensiveness.<br><br>3: The reasoning is moderately complete, but certain details are missing, requiring enhancement.<br><br>4: The reasoning is largely comprehensive, but a few minor details could be refined for better alignment with the question.<br><br>5 (good): The reasoning is comprehensive and includes all relevant information needed to fully justify the chosen answer. |
| Readability of Response: Assess whether the reasoning provided by the model is clear and easy to understand. | 1 (bad): The reasoning is highly difficult to read, full of grammatical errors, and lacks coherence and clarity. |
| | 2: The reasoning is somewhat difficult to read, and there are occasional grammatical errors. The coherence and clarity could be improved.
3: The reasoning is moderately easy to read, but there are noticeable grammatical errors and some parts lack coherence and clarity.
4: The reasoning is fairly easy to read, with only a few minor grammatical errors. Overall coherence and clarity are good, but there is room for improvement.
5 (good): The reasoning is easy to read, well-structured, and flows naturally. |

Table S3A: Confusion Matrix for Ophthalmology QA Extraction from MedMCQA Using Keyword Matching

| Keyword-based-retrieval | Actual ophthalmological question (n=6990) | Actual non-ophthalmological question (n=187,015) | Rate |
|---|---|---|---|
| **Predicted ophthalmological question (n=9,825)** | 5,016 | 4809 | True Positive Rate / Sensitivity: 0.718 |
| **Predicted non ophthalmological question (n=177,180)** | 1,974 | 175,206 | True Negative Rate / Specificity: 0.973 |

Table S3B: Confusion Matrix for Ophthalmology QA Extraction from MedMCQA Using Fine-Tuned PubMedBERT

| Finetuned PubMedBERT | Actual ophthalmological question (n=6990) | Actual non-ophthalmological question (n=187,015) | Rate |
|---|---|---|---|
| Predicted ophthalmological question (n=13,407) | 6,551 | 6,856 | True Positive Rate / Sensitivity: 0.937 |
| Predicted non ophthalmological question (n=173,598) | 439 | 173,159 | True Negative Rate / Specificity: 0.962 |

Table S4: Results from the Manual Quality Check and Grading Exercise

|  | No reasoning | Label 1 | Label 2 | Label 3 |
|---|---|---|---|---|
| **Number of items** | 40 | 92 | 382 | 386 |
| **Percentage (%)** | 4.4 | 10.2 | 42.4 | 42.9 |

Table S5: Performance of the models evaluated on the BELO dataset in terms of accuracy, completeness and comprehensiveness

| Metrics | GPT-4o | Llama-3-8B | Gemini 1.5 Pro |
|---|---|---|---|
| **Accuracy** | **4.91** | 4.72 | 4.81 |
| **Completeness** | 4.68 | 4.59 | **4.79** |
| **Readability** | **4.92** | 4.78 | 4.71 |

Appendix: Descriptions of BELO Dataset Development

1. Collection of medical datasets

We systematically reviewed and collected existing medical QA datasets (Figures 1 and 2). The search performed across February 2024, using multiple platforms, including Hugging Face, Google Scholar, PubMed Central, Kaggle, and Papers with Code. We used case-insensitive search terms such as "medical QA dataset", "medical NLP task", "QA dataset", "medical QA benchmark", "medical NLP benchmark", "ophthalmology exam questions", "ophthalmology MCQ", and "medical MCQ", without applying Boolean operators or wildcard searches.

2. Detailed Descriptions of the Datasets Used to Curate BELO

Basic and Clinical Science Course (BCSC)

The BCSC dataset consists of ophthalmology-related MCQs derived from the BCSC textbook, which is a comprehensive educational resource published by the American Academy of Ophthalmology (AAO). Each question offers four options, out of which only one is correct.

BioASQ Task B Phase B

The BioASQ dataset consists of biomedical research questions. Models are tasked to extract answers to questions about snippets from scientific literature given in the question. This dataset is expertly annotated with reasonings and has been updated annually since 2013 as part of the international BioASQ AI challenge. For our study, we used the questions from the 12th challenge which was released in 2024. Each question offers two options out of which only one is correct.

MedMCQA

This dataset consists of exam questions from two Indian affiliations, AIIMS (All India Institute of Medical Sciences) and NEET PG (National Eligibility cum Entrance Test for Post Graduate courses). These questions span 21 different medical subjects with subject-level labels. Questions vary from factual questions to diagnostic or treatment scenario questions based on patient cases. Each question offers four options out of which only one is correct.

MedQA

MedQA is a multilingual dataset featuring multiple-choice questions derived from medical board exams in the USA, Mainland China, and Taiwan. Our study focused on the English subset, which contains questions modeled after the United States

Medical Licensing Examination (USMLE) and created by medical examination experts. All questions are presented in a complex format based on patient case scenarios. Each question offers four options, out of which only one option is correct.

PubMedQA

PubMedQA consists of biomedical research questions. The dataset consists of 1000 question items with reasonings manually written by experts, 61.2k without reasonings and 211.3k with artificially generated reasonings. We extracted the QA items with reasonings written by experts to ensure that only high-quality MCQ items are used. Each question offers three options, out of which one is correct.

3. Detailed Descriptions of the Methodologies Used to Extract Ophthalmological QAs

3.1 Details of the Keyword Matching Methodology for Ophthalmological QA Extraction

We assembled a list of ophthalmology-specific terms (e.g. "glaucoma," "macula," "cataract") informed by terminology commonly found in ophthalmology board exams, peer-reviewed literature, and clinical documentation. Using a Python script, we scanned the "question" and "answer" fields of each QA pair; entries containing any keyword as a whole word were flagged as potentially ophthalmological. All matches were logged to generate an initial candidate set. To minimize false positives, we excluded ambiguous terms with high cross-specialty usage (e.g., "blind") and applied a secondary filter using keywords like "animal" to remove items referencing animal studies.

3.2 Finetuned PubMedBERT Model Methodology for Ophthalmological QA Extraction

For the Finetuned PubMedBERT model-based extraction, ophthalmology question detection was framed as a binary classification task. The PubMedBERT model was further finetuned with 100 ophthalmological and 100 non-ophthalmological questions from MedMCQA. MedMCQA was used due to the availability of subject-level labelled data.

4. Definitions of The Text-Generation Metrics Used in This Study

Five text-generation metrics were used in this study to evaluate model reasoning. The first text-generation metric used was the Recall-Oriented Understudy for Gisting Evaluation (ROUGE-L) which measures the longest common sequence (LCS) of words found between the model output and the BELO reference ground truth reasoning. The second metric used was the Metric for Evaluation of Translation with Explicit Ordering (METEOR) which calculates semantic similarity by measuring stemming, synonyms and exact word matches between the model output and the ground truth. It is calculated through a combination of precision and recall. Thirdly, BERTScore was used to measure semantic similarity between the model output and the ground truth using contextual embeddings from the Bidirectional Encoder Representations from Transformers (BERT) model. Furthermore, BARTScore was used to measure semantic similarity and fluency of the text compared to the ground truth by predicting the reference reasoning from the model output and vice versa, using the Bidirectional and Auto-Regressive Transformers (BART) model. Lastly, AlignScore was used to measure factual consistency through seminal and syntactic alignment between the model output and the ground truth.